%% file: main.tex
\documentclass{article}
\usepackage{iclr2027_conference,times}
\input{math_commands.tex}

\usepackage{hyperref}

\definecolor{linkblue}{RGB}{40,70,135}
\hypersetup{
    colorlinks=true,
    citecolor=linkblue,
    linkcolor=linkblue,
    urlcolor=linkblue,
    pdfpagelayout=OneColumn,
    pdfpagemode=UseNone,
    pdfstartview=FitH
}
\usepackage{url}
\usepackage{booktabs}
\usepackage{multirow}

\usepackage{amsmath,amssymb,amsthm}
\usepackage{xspace}
\usepackage{wrapfig}
\usepackage{algorithm}
\usepackage{algorithmic}
\usepackage{fontawesome5}
\newtheorem{proposition}{Proposition}
\newif\ifnotes \notestrue

\usepackage{graphicx}

\usepackage[table]{xcolor}

\definecolor{ctmgreen}{RGB}{223,240,216}
\definecolor{ctmred}{RGB}{248,215,218}
\definecolor{ctmgray}{RGB}{240,240,240}

\usepackage{tikz}
\definecolor{ktbar}{RGB}{128,128,128}
\definecolor{ktbg}{RGB}{242,242,242}
\newlength{\ktbarh}
\newsavebox{\ktbox}

\usepackage{placeins}
\usepackage{xspace}
\graphicspath{{figs/}}

\newcommand{\lk}{\texttt{<look>}\xspace}
\newcommand{\rdw}{\texttt{<read>}\xspace}

\title{Grounding on Request: Learning Modality Control in Vision-Language Models}

\begin{document}
\maketitle



\begin{abstract}
When an image and its caption disagree, can a user reliably control which one a vision-language model (VLM) follows? Accuracy alone cannot answer this question, since a model might give the correct answer from parametric knowledge without following the requested evidence. We therefore evaluate modality grounding control through conflict-preserving counterfactual interventions. Specifically, \textsc{Flip} measures whether an initially correct answer follows an edit to the requested modality, while \textsc{Stay} measures whether it remains stable when the competing modality is edited. To learn this control, we build on \textit{neologism learning} by introducing new words for visual and textual grounding, training only their embeddings while keeping the pretrained model fixed. We also introduce Counterfactual Twin-Margin (CTM) supervision, which uses paired interventions to teach how answer probabilities should change when the requested evidence changes. Across three VLMs on three existing benchmarks and our multi-attribute Ground-CF benchmark, we find that learned modality words improve control over natural-language instructions. Surprisingly to us, a single learned embedding per modality provides strong lightweight visual control and shows little average degradation on held-out values. LoRA remains strongest overall, although its visual advantage narrows under held-out evaluation. Moreover, counterfactual supervision improves control in several settings where standard supervision is weaker. Finally, we test our method on real-world news video, and find that visual control improves when the relevant evidence is made easier to access. Overall, our results show that reliable modality control requires testing not only whether an answer is correct, but whether it continues to follow edits to the modality of choice. 
\end{abstract}

\begin{center}
\faGithub\quad
\href{https://github.com/Maximeswd/Grounding-on-Request}
{\texttt{Maximeswd/Grounding-on-Request}}
\end{center}

\section{Introduction}
\label{sec:intro}

Vision-language models increasingly receive several modalities at the same time, including images, videos, and accompanying text \citep{bordes2024introductionvisionlanguagemodeling, fang2026unifiedvisionlanguagemodelsincomplete, multiventg2024, sanders2023multivent}. Often these sources agree, but what should happen when these modalities do not? In settings with cross-modal conflict \cite{zhang2025robust, zhu2024unravelingcrossmodalityknowledgeconflicts}, the appropriate answer can depend on which modality the user wants the model to follow \cite{catapang-2026-image,hua2025visionlanguagemodelsprocessconflicting}. In practice, however, VLMs can exhibit \emph{modality dominance}, favoring one modality (often text) even when explicitly asked to use another \citep{ghatkesar2025beyond,hua2025visionlanguagemodelsprocessconflicting,Jia_2025_CVPR}. Previous work has therefore focused largely on either strengthening visual grounding or reducing the influence of dominant textual information \citep{ghatkesar2025beyond,shi2026vlmsseeingjustsaying,wu2025languageoverrulesrevealingtext}. Rather than pushing the model permanently toward one modality, we seek to understand how we can allow the user to choose either modality on demand.

\begin{figure}[t]
\centering
\includegraphics[width=\linewidth]{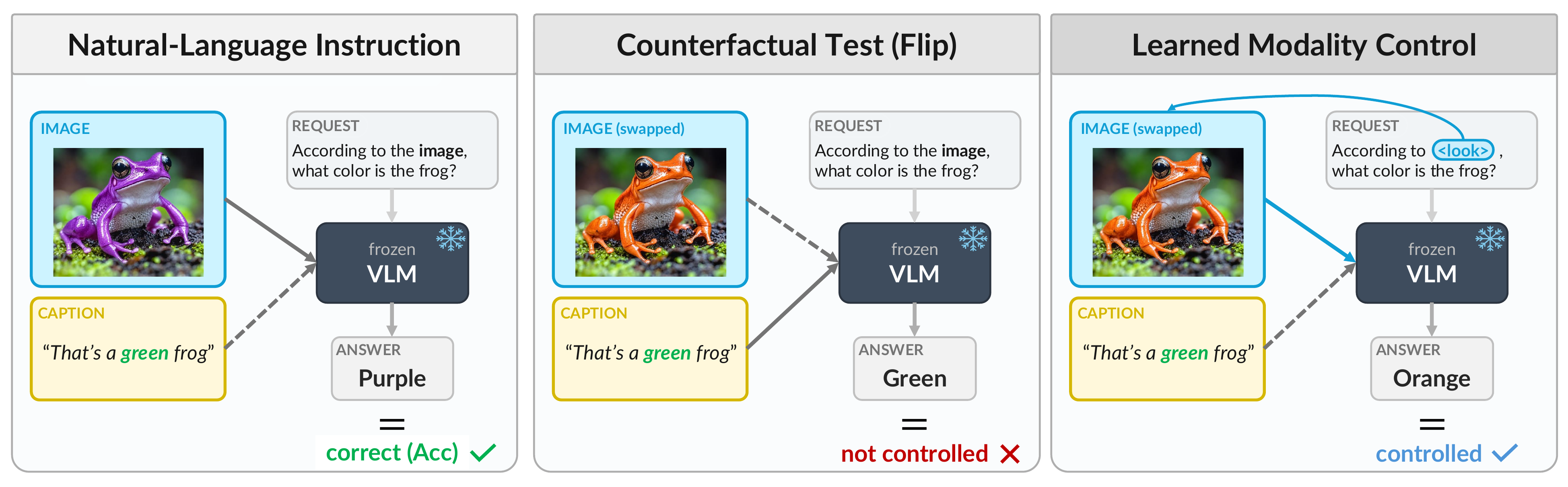}
\caption{\textbf{Accuracy misses modality grounding control.} Left: the image shows a purple frog, the caption says green, and the model correctly answers purple under an image-grounding request. Middle: after the image changes to orange, answering green fails to follow the requested modality. Right: the desired answer is orange, shown with the learned visual grounding word \lk{}.}
\label{fig:assumption}
\end{figure}

Crucially, measuring this modality control requires more than just checking accuracy. Suppose an image shows a purple frog, but the caption describes it as green (\autoref{fig:assumption}). If we ask the model to look at the image and it answers \emph{purple}, standard evaluation marks this as correct. However, correctness alone does not establish that the answer reliably follows the image. To test this, we change only the requested modality (the image) so that the frog becomes orange. An image-grounded answer should then change to \emph{orange}. On the other hand, if we only change the competing caption, an image-grounded answer should remain unchanged. We therefore evaluate modality grounding through paired counterfactual interventions. Specifically, \textsc{Flip} measures whether an initially correct answer follows an edit to the requested modality, while \textsc{Stay} measures whether it remains stable when the competing modality changes.


Our work aims to bridge the gap between measuring modality control and explicitly learning it. In particular, we expand on \emph{neologism learning} \citep{hewitt2025position,hewitt2025neologism}, which introduces new words with new word embeddings. These words are placed directly in natural language, while only their embeddings are trained and the pretrained model remains fixed. We introduce \lk{} for visual grounding and \rdw{} for textual grounding. Since standard supervised fine-tuning (SFT) optimizes examples independently, it does not explicitly teach how answer preferences should move when the requested evidence changes. To address this, we introduce the \emph{Counterfactual Twin-Margin} (CTM) objective, which uses paired interventions so that the original answer becomes less likely and the replacement answer more likely after an edit to the requested modality.

To study this systematically, we evaluate three VLMs on three existing benchmarks and construct \textbf{Ground-CF}, a controlled benchmark with image–text interventions over color, shape, count, and spatial relations. We compare learned modality words with natural-language instructions \citep{shi2026vlmsseeingjustsaying}, parameter-matched prompt tuning \citep{lester2021prompt}, and higher-capacity LoRA adaptation \citep{hu2022iclr-lora}. We find that natural-language instructions provide substantially weaker modality control, particularly for visual requests. Surprisingly to us, a single learned embedding per modality already provides strong lightweight visual control and remains stable on average across held-out attribute values. LoRA achieves the strongest absolute control, while its visual advantage becomes smaller on the held-out evaluations. Moreover, counterfactual supervision improves control in several settings where standard supervision is weaker. Finally, we take it one step further and evaluate these methods on real-world news videos using MultiVENT-G \citep{multiventg2024}. Here, Neologisms again provide the strongest lightweight visual control, despite using substantially fewer trainable parameters than LoRA, and improve considerably when the relevant evidence is made easier to access.

Overall, our work contributes: (1) a conflict-preserving counterfactual formulation of modality grounding control using \textsc{Flip} and \textsc{Stay}, together with Ground-CF; (2) evidence that accuracy and natural-language instructions substantially overstate reliable visual control; (3) adapting neologism learning for user-selectable modality grounding, together with CTM-based counterfactual supervision; and (4) evidence that lightweight learned controls are effective on held-out attributes and extend to realistic news-video frames, where access to relevant visual evidence becomes increasingly important.


\section{Related Work}
\label{sec:related}

\noindent\textbf{Modality conflict and grounding.}
There has been substantial progress in understanding how VLMs combine visual and textual information, particularly when the two modalities disagree \cite{liu2025visual,luo2025visionlanguage,Deng_2025_CVPR}. Prior work finds that models often rely unevenly on their multimodal inputs, with language priors frequently dominating visual evidence \citep{ghatkesar2025beyond,catapang-2026-image,ghosh2026whichsource,li2026signpost}. This has motivated methods that strengthen visual grounding or reduce the influence of dominant textual information, including contrastive region guidance \citep{crg2024}, specialized visual perception tokens \citep{vpt2025}, active visual querying \citep{vithinker2026}, perception-focused rewards \citep{badseeing2026}, information-flow interventions \citep{routegating2026}, and explicit user instructions \citep{shi2026vlmsseeingjustsaying,wu2025languageoverrulesrevealingtext}. These works largely study which modality dominates and how conflict leads to hallucinations. We approach the problem from a different angle. Rather than deciding which modality should generally dominate, we ask: given a conflict, can the user select \textit{either} modality at inference time, and will that choice remain reliable under intervention?

\noindent\textbf{Counterfactual dependence and attribution.}
Previous work has explored empirically whether model outputs are supported by, or depend on, particular evidence \citep{mavis2026,tracer2026,wallat2024faithfulness}. Since evidence that supports an answer does not establish that the answer depended on it, counterfactual interventions provide a stronger behavioral test by changing the evidence and measuring whether the prediction changes accordingly \citep{kamahi2024counterfactuals}. VisualSwap, for example, measures whether VLMs respond when visual evidence is replaced \citep{shi2026vlmsseeingjustsaying}. Our work builds on this idea, but we test both sides of user-selected modality control. Specifically, we measure whether predictions follow changes to the requested modality and remain stable when only the competing modality changes. We also go one step further and use these paired interventions during training rather than only for evaluation.

\noindent\textbf{Steering, neologisms, and training objectives.}
There has also been substantial progress on controlling model behavior, moving from interventions on internal representations \citep{huben2024sparse,rimsky-etal-2024-steering,tan2024analysing,liu2024incontextvectorsmakingcontext, hua2026sourcemodality}, to lightweight adaptations such as prompt tuning \citep{lester2021prompt,lilang2021prefix}, or LoRA \citep{hu2022iclr-lora}. In particular, we expand on \emph{neologism learning}, where new words are introduced with new word embeddings while the pretrained model remains fixed \citep{hewitt2025position,hewitt2025neologism}. These words have previously been used to learn concepts and behaviors such as response style or simple skills \citep{hewitt2025neologism,berthon2026skill}. We instead use them as explicit user-selectable modality controls. Standard SFT can train on both members of a counterfactual pair, but optimizes each context independently. In contrast, CTM directly couples the same candidate answers across paired evidence edits. This allows us to train the directional behavior tested by our evaluation: when the requested modality changes, the old answer should become less likely while the replacement becomes more likely.

\section{Measuring and Learning Modality Grounding Control}
\label{sec:criterion}

As described in the previous section, a correct answer under a modality request does not establish that the answer reliably follows the requested evidence. To quantify the effectiveness of modality grounding control, we therefore ask how predictions respond to edits of the modality inputs. When the requested modality changes, the answer should follow it. On the other hand, when only the competing modality changes, the answer should remain the same. We first formalize these two behaviors through paired counterfactual interventions, and then use the same interventions to learn stronger modality controls.

\subsection{Measuring Modality Grounding Control}

A modality grounding request $r_s$, such as a natural-language instruction or learned word, selects which source $s$ should ground the answer. We investigate three behaviors: the model should answer correctly from the requested modality, follow it when it changes, and remain stable when only the competing modality changes.

Let $C=(q,s_1,\ldots,s_M)$ contain a question $q$ and $M$ evidence sources, where the requested source $s$ supports answer $v_s$. Since each benchmark provides a fixed set of valid candidate answers, we evaluate the model directly through their sequence log-likelihoods rather than relying on open generation or an external judge. For candidate $y=(y_1,\ldots,y_T)$, we therefore score:

\begin{equation}
    m_s(y\mid C)
    =
    \sum_{t=1}^{T}\log p(y_t\mid C,r_s,y_{<t}),
    \qquad
    \hat y_s(C)
    =
    \arg\max_{y\in\mathcal Y} m_s(y\mid C).
\label{eq:candidate_scoring}
\end{equation}

Here, $m_s(y\mid C)$ sums the token log-probabilities of candidate $y$ under grounding request $r_s$, and $\hat y_s(C)$ is the selected candidate.

\paragraph{Accuracy.}
Accuracy measures whether the model returns the value supported by the requested modality, $ A_s=\Pr[\hat y_s(C)=v_s]$. This measures correctness on the original input, but not whether the answer continues to follow that modality when the evidence changes.

\paragraph{Flip and Stay.}
To measure this control directly, we construct two counterfactual twins. For \textsc{Flip}, $C_s^\prime$ changes only the requested modality, replacing $v_s$ with $v_s^\prime$. For \textsc{Stay}, $C_u^\prime$ changes only the competing modality $u\neq s$, while $v_s$ remains unchanged. We measure:

\begin{equation}
\textsc{Flip}_s
=
\Pr\!\left[
\hat y_s(C)=v_s
\wedge
\hat y_s(C_s')=v_s'
\right],
\qquad
\textsc{Stay}_{s,u}
=
\Pr\!\left[
\hat y_s(C)=v_s
\wedge
\hat y_s(C_u')=v_s
\right].
\label{eq:flip_stay}
\end{equation}

Both metrics are conjunctive at the item level. A successful \textsc{Flip} requires the answer to remain correct before and after the requested-modality edit, while a successful \textsc{Stay} requires it to remain correct after an edit to the competing modality. Thus, $\textsc{Flip}s\leq A_s$ and $\textsc{Stay}{s,u}\leq A_s$. High Accuracy with lower \textsc{Flip} or \textsc{Stay} therefore exposes initially correct answers that do not reliably behave according to the requested source.

Previous work has used interventions to test whether VLM predictions follow changes in visual evidence \citep{shi2026vlmsseeingjustsaying,hua2025visionlanguagemodelsprocessconflicting,zhu2024unravelingcrossmodalityknowledgeconflicts}. We go one step further and test both sides of user-selected modality control: whether predictions follow changes to the requested modality and remain stable when only the competing modality changes.

\subsection{Learning Modality Control with Counterfactual Supervision}

In this section, we draw from and extend neologism learning \citep{hewitt2025position,hewitt2025neologism} to learn modality-specific controls. At a high level, neologism learning keeps the pretrained model fixed, expands its vocabulary and embedding matrix with new words, and optimizes only these new embeddings on examples that exemplify a desired concept or behavior. We introduce \lk{} for visual grounding and \rdw{} for textual grounding, represented by $L$ newly added tokens per word. These words are placed directly in natural language, while only their embeddings are trained. The added token IDs are masked from the output vocabulary so they can solely be used as control inputs but not generated as answers. Since standard SFT optimizes examples independently, it does not explicitly constrain how answer preferences should change across a counterfactual pair. Our work aims to bridge this gap through the \emph{Counterfactual Twin-Margin} (CTM) objective.

Consider a pair $(C,C_s^\prime)$ in which the requested modality supports answer $v$ before the edit and $v^\prime$ afterwards. We seek to make the old answer less likely after the edit and the replacement answer more likely. Specifically, we define:

\begin{equation}
\Delta_{\rm old}
=
m_s(v\mid C)-m_s(v\mid C_s'),
\qquad
\Delta_{\rm new}
=
m_s(v'\mid C_s')-m_s(v'\mid C).
\label{eq:ctm_shifts}
\end{equation}

Thus, positive values of $\Delta_{\rm old}$ and $\Delta_{\rm new}$ indicate that both answers move in the desired direction across the intervention.

We encourage these shifts with a logistic loss:

\begin{equation}
\mathcal L_{\rm CTM}
=
-\log\sigma(\gamma\Delta_{\rm old})
-\log\sigma(\gamma\Delta_{\rm new}),
\qquad \gamma>0.
\label{eq:ctm}
\end{equation}

We then train with $\mathcal L=\mathcal L_{\rm SFT}+\mathcal L_{\rm CTM}$.

\begin{proposition}[Two-sided margins move relative preference]
\label{prop:ctm}
Let $D(C)=m_s(v\mid C)-m_s(v'\mid C)$ denote the log-probability ratio of the original answer to the replacement answer. Then: $D(C)-D(C_s')=\Delta_{\rm old}+\Delta_{\rm new}.$
Consequently, if both CTM margins are positive, then $D(C)>D(C_s')$.
\end{proposition}

Thus, CTM moves the relative preference toward the replacement answer after the requested modality changes. Importantly, this does not guarantee a successful \textsc{Flip}: the preference may not cross the decision boundary, or another candidate may still score higher. Moreover, CTM does not explicitly train invariance to edits of the competing modality. Interestingly, as we show later, this distinction matters empirically. Hence, we continue to evaluate \textsc{Flip} and \textsc{Stay} separately. \autoref{app:proofs} provides the derivation and counterexamples.

\section{Experimental Setup}
\label{sec:setup}

We test modality grounding control across models, benchmarks, and adaptation methods. Our main experiments use controlled image–text conflicts, where we can intervene on one modality while keeping the other fixed. Finally, we go one step further and evaluate the same control methods on more realistic news-video frames.

\subsection{Benchmarks}
We evaluate four controlled image–text benchmarks. \textbf{Visual-Counterfactual} \citep{golovanevsky-etal-2025-pixels} tests colour changes while preserving object identity. \textbf{ColorSwap} \citep{burapacheep2024colorswap} tests colour–object binding by reassigning colours between objects, while \textbf{COCO-CF} \citep{le2023cococounterfactuals} contains controlled entity substitutions in natural image–text pairs. We also introduce \textbf{Ground-CF}, inspired by CLEVR \citep{johnson2017clevr}, which extends these interventions across colour, shape, count, and spatial relations. For all primary evaluations, we construct \emph{conflict-preserving} counterfactual twins. Specifically, when one modality is edited, its replacement value differs from both its original value and the value supported by the competing modality. This allows us to test whether the model continues to follow the requested source while contradictory evidence remains present. Full construction details and a comparison with agreement-resolving interventions are provided in \autoref{app:protocol}. Finally, \textbf{MultiVENT-G} \citep{sanders2023multivent,multiventg2024} provides real-world news-video frames, text, and human-annotated visual evidence. We use it as a separate application setting to test modality control under substantially more complex visual inputs. Dataset adaptations and evaluation details are given in \autoref{app:mvg}.

\noindent\textbf{Models and configurations.}
We evaluate Gemma-3-4B, Qwen3.5-4B, and Qwen3.5-2B \citep{gemmateam2025gemma3technicalreport,yang2025qwen3technicalreport}. We compare learned neologisms with the frozen base model, natural-language grounding instructions \citep{shi2026vlmsseeingjustsaying}, parameter-matched prompt tuning \citep{lester2021prompt}, and higher-capacity LoRA adaptation \citep{hu2022iclr-lora}. Neologism learning and prompt tuning use $L=1$ learned embedding per request, unless stated otherwise. Prompt tuning prepends the learned vector to the input, while neologism learning adds a new vocabulary token that is used directly inside the natural-language grounding request. LoRA uses rank $r=16$ and one specialist adapter per requested modality, providing a higher-capacity, source-specific baseline. All learned methods use SFT+CTM supervision in the main comparisons. We evaluate with Accuracy, \textsc{Flip}, and \textsc{Stay} as defined in \autoref{sec:criterion}. Full prompts, optimization settings, candidate sets, checkpoint selection, and intervention coverage are provided in \autoref{app:protocol}.

\section{Analytical Results}

We first study modality grounding under controlled image--text conflicts. We begin by asking which modality the unadapted models follow when no grounding request is provided. We then test whether natural-language requests reliably change this behavior, before asking whether modality control can instead be learned.

\noindent\textbf{Models have different default modality preferences.}
Before asking how to control modality grounding, we first examine what models do when image and text disagree. A model may select the value supported by the image, the text, or neither. \autoref{tab:modality-dominance} shows that modality preference varies substantially across both models and benchmarks. On COCO-CF, all three models favor textual evidence, whereas on Ground-CF and ColorSwap both Qwen models favor the image. VCF shows further variation across models.

\begin{table}[htbp]
\centering
\setlength{\tabcolsep}{4.5pt}
\begin{tabular}{lcccccc}
\toprule
& \multicolumn{2}{c}{Gemma-3-4B}
& \multicolumn{2}{c}{Qwen3.5-4B}
& \multicolumn{2}{c}{Qwen3.5-2B} \\
\cmidrule(lr){2-3}
\cmidrule(lr){4-5}
\cmidrule(lr){6-7}
Benchmark
& Image & Text
& Image & Text
& Image & Text \\
\midrule
VCF
& .40 & \textbf{.60}
& .50 & .50
& \textbf{.67} & .33 \\
ColorSwap
& .43 & \textbf{.57}
& \textbf{.68} & .32
& \textbf{.74} & .26 \\
COCO-CF
& .32 & \textbf{.68}
& .16 & \textbf{.84}
& .36 & \textbf{.64} \\
Ground-CF
& .43 & \textbf{.50}
& \textbf{.70} & .23
& \textbf{.62} & .24 \\
\bottomrule
\end{tabular}
\caption{\textbf{Default modality preference depends on both model and benchmark.} We report how often each base model selects the value supported by the image or text on conflicting examples. The larger value is shown in bold.}
\label{tab:modality-dominance}
\end{table}

This variation is important for modality control, because if the same modality dominated across models and benchmarks, one could potentially address the problem by strengthening the weaker source or suppressing the dominant one. Instead, we find that there is no single modality that should generally be amplified or suppressed. A reliable modality grounding control method should therefore allow the user to select either modality on demand, irrespective of the base model's default preference.

\subsection{Accuracy Does Not Measure Reliable Modality Grounding Control}
\label{sec:diagnosis}

We next ask whether natural-language instructions already provide this control. Standard multimodal evaluation focuses on whether a model gives the correct answer (Accuracy). We approach the problem from a different angle, asking whether that answer continues to follow the same modality when the evidence changes. If the requested image changes, the answer should change accordingly. On the other hand, if only the competing caption changes, the answer should remain the same. \textsc{Flip} and \textsc{Stay} measure these two behaviors directly.

\begin{figure}[htbp]
\centering
\includegraphics[width=\textwidth]{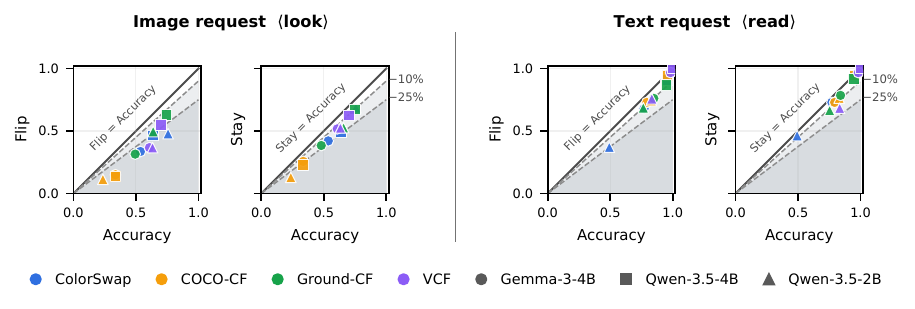}
\caption{\textbf{Accuracy overstates reliable modality control.} Results are shown for natural-language image requests (left) and text requests (right). If every initially correct answer behaves appropriately under intervention, FLIP or STAY equals Accuracy. Points below the diagonal therefore identify correct predictions that fail the corresponding counterfactual test.}
\label{fig:accgrounding}
\end{figure}

\autoref{fig:accgrounding} shows that natural-language instructions do not provide reliable modality control, particularly for visual grounding requests. Across all $12$ model--benchmark settings, every image request loses at least a $10\%$ of its initially correct answers under the requested-modality intervention, and in $9$ settings at least $25\%$ fail to follow the image edit (\textsc{Flip}). For example, Gemma-3-4B reaches $.336$ Accuracy on COCO-CF under an image instruction, but drops to $.150$ under \textsc{Flip}. On the other hand, text requests are considerably more reliable. Only $5$ of $12$ settings lose at least $10\%$, and none loses $25\%$ under \textsc{Flip}. We find the same asymmetry for \textsc{Stay}, with $11$ of $12$ image-request settings showing at least a $10\%$ drop, compared with only $2$ of $12$ in the text-request settings. 


Thus, correctness and control are not equivalent. And overall, the two analyses motivate both parts of our approach. Modality dominance varies across models and benchmarks \autoref{tab:modality-dominance}, so a general solution cannot simply steer every model toward the same modality. Moreover, natural-language instructions do not reliably provide user-selectable modality grounding control, and Accuracy alone would hide these failures. We therefore want to investigate whether a learned modality grounding request can provide more reliable control over either modality, and evaluate this control through \textsc{Flip} and \textsc{Stay}.

\subsection{Learning Reliable Modality Grounding Control}
\label{sec:control}
Rather than pushing the model permanently toward vision or text, we seek a lightweight modality control that allows either modality to be selected on demand. We compare neologisms with parameter-matched prompt tuning and a substantially higher-capacity, source-specific LoRA baseline. We evaluate both in-distribution (ID) and on held-out values or concepts.

\autoref{tab:control_id_ood} reveals a clear difference between the two modalities. Textual control is comparatively easy, since all learned methods approach the ceiling, with prompt tuning slightly stronger than neologisms and LoRA strongest overall. Visual control, on the other hand, remains substantially harder. Here, neologisms provide the strongest lightweight control, outperforming prompt tuning on both \textsc{Flip} and \textsc{Stay} in ID and OOD. This aggregate includes one Gemma-3-4B COCO-CF prompt-tuning failure, while the remaining cells are substantially closer; full per-cell results are reported in \autoref{app:fulltables}. We therefore interpret the aggregate result as evidence that neologisms provide competitive lightweight visual control across the full grid while avoiding the benchmark-specific collapse observed for prompt tuning.

Interestingly, neologism visual control remains almost unchanged on held-out values: \textsc{Flip} decreases by only $0.7\%$ and \textsc{Stay} by $1.6\%$. Prompt tuning is similarly stable on average, while LoRA remains strongest in absolute terms but loses more of its visual advantage, with \textsc{Flip} decreasing by $9.4\%$ and \textsc{Stay} by $8.6\%$. Overall, a single learned embedding per request provides strong visual grounding control with very few trainable parameters.

\begin{table}[htbp]
\centering
\small
\setlength{\tabcolsep}{5.2pt}
\setlength{\tabcolsep}{3.2pt}\footnotesize
\begin{tabular}{lcccccccccccc}
\toprule
&
\multicolumn{6}{c}{Visual request \lk{}}
&
\multicolumn{6}{c}{Text request \rdw{}}
\\
\cmidrule(lr){2-7}
\cmidrule(lr){8-13}
Method
& \multicolumn{3}{c}{ID}
& \multicolumn{3}{c}{OOD}
& \multicolumn{3}{c}{ID}
& \multicolumn{3}{c}{OOD}
\\
& A & F & S & A & F & S
& A & F & S & A & F & S
\\
\midrule
Base model & .398 & .208 & .294 & .419 & .251 & .324 & .546 & .422 & .435 & .513 & .417 & .405 \\
Instruction & .555 & .367 & .455 & .573 & .401 & .486 & .848 & .784 & .799 & .825 & .762 & .784 \\
Prompt-tuning$^\dagger$ & .742 & .630 & .683 & .734 & .630 & .683 & \underline{.979} & \underline{.968} & \underline{.976} & \underline{.981} & \underline{.971} & \underline{.979} \\
Neologism & \underline{.800} & \underline{.664} & \underline{.733} & \underline{.779} & \underline{.657} & \underline{.717} & .940 & .917 & .930 & .921 & .902 & .912 \\
LoRA & \textbf{.911} & \textbf{.800} & \textbf{.869} & \textbf{.833} & \textbf{.706} & \textbf{.783} & \textbf{.999} & \textbf{.999} & \textbf{.999} & \textbf{.998} & \textbf{.997} & \textbf{.998} \\
\bottomrule
\end{tabular}
\caption{\textbf{Learned modality requests substantially improve control, while visual grounding remains harder.} Scores are averaged across three models and four benchmarks. We report \textsc{Flip} (F) and \textsc{Stay} (S) for ID and held-out evaluation. $^\dagger$The prompt-tuning visual average includes one collapsed Gemma-3-4B COCO-CF cell (image Accuracy $.050$), the full per-cell results are reported in \autoref{app:fulltables}. Bold denotes the best method and underlining the runner-up.}
\label{tab:control_id_ood}
\end{table}

\begin{wrapfigure}{r}{0.4\textwidth}
\centering
\includegraphics[width=0.95\linewidth]{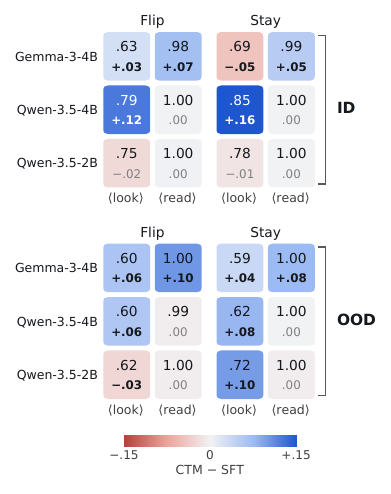}
\caption{\textbf{CTM helps most where SFT is weaker.} Ground-CF results for neologisms trained with SFT or CTM; cells show CTM and CTM$-$SFT (bold: McNemar $p<.05$).}
\label{fig:ctm_sft}
\end{wrapfigure}

The previous comparison fixes SFT+CTM across learned methods. We now isolate the role of counterfactual supervision within neologism learning by comparing the same $L{=}1$ word trained with SFT or SFT+CTM. Since CTM also introduces paired counterfactual contexts, we additionally compare against paired SFT with the same twins but no cross-context coupling. Interestingly, exposure to the twins alone improves caption control but not visual control, whereas CTM improves visual \textsc{Flip} beyond paired SFT in both ID and held-out evaluation. Thus, the visual-control gain cannot be explained by counterfactual exposure alone and instead depends on coupling the paired contexts. Full results are reported in \autoref{app:objective}.

\autoref{fig:ctm_sft} shows that CTM helps most where SFT is weaker. In ID evaluation, the clearest gain appears for Qwen3.5-4B visual grounding, where \textsc{Flip}/\textsc{Stay} increase by $+.12/+.16$. On held-out values, improvements are broader, where Gemma-3-4B improves across both requests, Qwen3.5-4B again improves on visual control, and Qwen3.5-2B gains $+.08$ in \textsc{Stay}. The effect is not uniform across every model and metric, with complete results reported in \autoref{app:objective}.

Interestingly, explicitly adding \textsc{Stay} supervision does not consistently improve control and can reduce \textsc{Flip}. On the other hand, CTM alone improves \textsc{Stay} in several settings despite never optimizing it directly. This suggests that learning to follow the requested modality can also induce selectivity toward competing evidence, while directly optimizing both behaviors may introduce a sensitivity–invariance trade-off. Overall, paired counterfactual supervision is most useful where standard supervision is weaker, particularly on held-out values.

\paragraph{Can the model verbalize what the learned words mean?} Since neologisms are actual new vocabulary words, we can ask the model directly what they mean. This is less natural for prompt tuning or LoRA, where the control is stored in learned vectors or adapter weights rather than in a word. As shown in \autoref{fig:meta_words}, models can often tell that \lk{} and \rdw{} are different, but are substantially less reliable at saying which modality each word refers to. On the other hand, these same words still strongly control which modality the model follows under intervention. This suggests that the learned words can control behavior without the model being able to clearly explain or self-verbalize their meaning. Since these probes use $L{=}1$ Ground-CF neologisms, stronger controls or additional learned tokens may make their meaning easier for the model to verbalize.

\vspace{1em}

\begin{figure}[htbp]
\centering
\begin{minipage}[t]{0.60\linewidth}
\small
\raggedright
\textbf{Metalinguistic probes}\\[8pt]
\textbf{1. Distinguishing the learned words}\\[-1pt]
{\color{violet}\texttt{Are \lk{} and \rdw{} the same or different?}}\\
{\color{purple}\hspace{1em}Gemma-3-4B: \emph{different}}\\[5pt]

\textbf{2. Identifying the source}\\[-1pt]
{\color{violet}\texttt{What does \lk{} ask you to use?}}\\
{\color{violet}\texttt{A. the image \quad B. the caption}}\\
{\color{purple}\hspace{1em}Gemma-3-4B: \emph{B. the caption}}
\end{minipage}
\hspace{0.02\linewidth}
\begin{minipage}[t]{0.34\linewidth}
\vspace{1.6em}
\centering
\scriptsize
\setlength{\tabcolsep}{3pt}
\begin{tabular}{@{}lcc@{}}
\toprule
& \shortstack{Distinguish\\words}
& \shortstack{Identify\\source} \\
\midrule
Gemma-3-4B & .750 & .500 \\
Qwen3.5-2B & .625 & .625 \\
\bottomrule
\end{tabular}
\end{minipage}
\caption{\textbf{Behavioral control does not imply self-verbalizable meaning.}
Models often distinguish the learned words, but are substantially less reliable at identifying the modality each word denotes.}
\label{fig:meta_words}
\end{figure}


\section{Applied Results}

\subsection{Modality Grounding Control on Real-World News Videos}
\label{sec:video}

Controlled benchmarks let us isolate whether a model follows the requested modality, but real-world images are harder because the relevant evidence may be only a small part of a complex scene. We therefore also evaluate on MultiVENT-G \citep{multiventg2024}, a news-video dataset with human-annotated visual evidence for event roles \citep{sanders2023multivent}. We construct counterfactual pairs by replacing a text role value with another value from the same class, or by selecting another frame from the same video in which that role is linked to a different visual entity. Unlike our controlled benchmarks, these interventions are not perfectly isolated, since changing the frame can also change other parts of the scene. Since this setting is more complex, we use $L{=}5$ learned embedding rows per request for neologisms and prompt tuning; LoRA remains rank $16$, and all learned methods are trained on MultiVENT-G with SFT+CTM. Full details are given in \autoref{app:mvg}.

\begin{table*}[htbp]
\centering
\small
\setlength{\tabcolsep}{6pt}
\begin{tabular}{lcc|cc|cc}
\toprule
&
\multicolumn{2}{c}{Gemma-3-4B}
&
\multicolumn{2}{c}{Qwen3.5-4B}
&
\multicolumn{2}{c}{Qwen3.5-2B}
\\
\cmidrule(lr){2-3}
\cmidrule(lr){4-5}
\cmidrule(lr){6-7}
Method
& $\langle$\texttt{look}$\rangle$
& $\langle$\texttt{read}$\rangle$
& $\langle$\texttt{look}$\rangle$
& $\langle$\texttt{read}$\rangle$
& $\langle$\texttt{look}$\rangle$
& $\langle$\texttt{read}$\rangle$
\\
\midrule
Base model
& $.214 \rightarrow .277$ & $.657$
& $.145 \rightarrow .202$ & $.693$
& $.087 \rightarrow .127$ & $.670$
\\

Instruction
& $.191 \rightarrow .266$ & $.727$
& $.162 \rightarrow .364$ & $.820$
& $.110 \rightarrow .173$ & $.760$
\\

Prompt-tuning
& $.173 \rightarrow \underline{.249}$ & $\underline{.870}$
& $.156 \rightarrow .347$ & $.807$
& $.133 \rightarrow .214$ & $\underline{.863}$
\\

Neologism
& $.104 \rightarrow .220$ & $.863$
& $.202 \rightarrow \underline{.486}$ & $\underline{.903}$
& $.145 \rightarrow \underline{.295}$ & $.850$
\\

LoRA
& $.191 \rightarrow \mathbf{.416}$ & $\mathbf{.983}$
& $.376 \rightarrow \mathbf{.595}$ & $\mathbf{.993}$
& $.353 \rightarrow \mathbf{.538}$ & $\mathbf{.997}$
\\
\bottomrule
\end{tabular}

\caption{
\textbf{Making the relevant visual evidence easier to find improves visual control on news frames.}
We report \textsc{Flip} on MultiVENT-G. For $\langle$\texttt{look}$\rangle$, scores compare the full frame with a crop centered on the human-annotated evidence region (Full $\rightarrow$ Crop), making the relevant visual information a larger part of the input. Crops are used only at evaluation time. For $\langle$\texttt{read}$\rangle$, we report full-frame performance. Visual pairs have $n{=}173$ and description pairs $n{=}300$ per model. Bold denotes the best learned method and underlining the runner-up.
}
\label{tab:mvg_crop}
\end{table*}

\autoref{tab:mvg_crop} shows that textual control remains strong, while full-frame visual control is considerably harder. Most $\langle$\texttt{look}$\rangle$ \textsc{Flip} scores are near or below the $.25$ chance level, even though per-frame accuracy remains between $.50$ and $.66$. Thus, the models can often answer individual frames correctly without reliably changing their answer when the visual evidence changes. Among the lightweight methods, neologisms are particularly strong on the Qwen models. After cropping, they outperform parameter-matched prompt tuning on both Qwen3.5-4B ($.486$ vs.\ $.347$) and Qwen3.5-2B ($.295$ vs.\ $.214$), with the largest gain on the larger Qwen3.5-4B model. Gemma-3-4B is the exception, where prompt tuning remains slightly stronger. This suggests that the benefit of the learned visual word depends in part on the underlying model's ability to use the available visual evidence.

\noindent\textbf{Evidence access limits visual modality control.}
Since MultiVENT-G provides human-annotated evidence boxes, we can test whether part of the remaining difficulty comes from finding the relevant visual evidence. Cropping around the annotated region improves visual \textsc{Flip} for every method and model, without retraining. Neologisms improve from $.104$ to $.220$ on Gemma-3-4B, from $.202$ to $.486$ on Qwen3.5-4B, and from $.145$ to $.295$ on Qwen3.5-2B. This is consistent with the broader pattern in our controlled experiments. Neologisms are competitive with prompt tuning on average, but their visual advantage becomes clearer in harder visual settings such as COCO-CF and on  MultiVENT-G. Overall, these results separate two sources of difficulty. A control mechanism may successfully specify \textit{which modality to use}, while the underlying VLM can still fail to identify \textit{where the relevant evidence is located} within that modality. Cropping does not solve evidence localization, as it supplies the relevant region, but the resulting improvement shows that evidence access accounts for part of the remaining visual-control gap.


\section{Conclusion}
\label{sec:conclusion}

We asked whether a user can reliably control which modality a VLM follows when image and text disagree. Our results show that correctness alone does not establish this control. We therefore introduce \textsc{Flip} and \textsc{Stay}, which test whether predictions follow edits to the requested modality while remaining stable to edits of competing evidence. Using these interventions, we find that modality control can be learned with very few parameters. Neologisms provide strong lightweight control, remain competitive with parameter-matched prompt tuning overall, and are particularly robust in harder visual settings. While LoRA remains strongest overall, its visual control degrades more under held-out evaluation, while using $5{,}300$--$11{,}600\times$ more trainable parameters than $L{=}1$ neologisms and prompt tuning. Moreover, CTM improves visual control beyond exposure to counterfactual examples alone in several settings. On MultiVENT-G, visual control improves substantially when the relevant evidence is made easier to find, showing that modality selection and evidence access are distinct grounding challenges. Overall, reliable modality control should be evaluated by how predictions respond when the underlying evidence changes, rather than through correctness alone.

\paragraph{Limitations.}
Our work establishes modality grounding control through behavioral interventions, rather than explaining the internal mechanisms that produce this behavior. Understanding what the learned neologism tokens encode, how they steer the model toward one source over another, and how this changes as their capacity increases are therefore interesting directions for future work. Moreover, we evaluate models using fixed answer choices and held-out values from the same type of task, which allows precise measurement of modality control but does not test arbitrary distribution shifts or fully open-ended generation. MultiVENT-G moves toward more realistic scenes, although its frame interventions are less controlled and the crop analysis relies on human-annotated evidence. Finally, future work should extend these findings to larger models, open-ended generation tasks, and other modalities such as audio.

\subsubsection*{AI use statement}
Language models assisted with writing, literature discovery, research ideation, experiment orchestration, code development, and data generation. In the MultiVENT-G experiments, Fable 5 generated claims from entity--role pairs used to construct dataset captions. The authors verified the generated data, experimental records, claims, citations, and mathematical arguments, and remain responsible for all reported results.

\section*{Acknowledgments}
This research was supported by project VI.Vidi.223.166 of the NWO Talent Programme which is (partly) financed by the Dutch Research Council (NWO).
We acknowledge the Dutch Research Council for awarding this project access to the LUMI supercomputer, owned by the EuroHPC Joint Undertaking, hosted by CSC (Finland) and the LUMI consortium through project NWO-2025.040.


\bibliography{references}
\bibliographystyle{iclr2027_conference}


\clearpage
\appendix

\section{Data and Evaluation Protocol}
\label{app:protocol}\label{app:datasets}

\paragraph{Benchmarks.}
We evaluate modality grounding control on Visual-Counterfactual (VCF), ColorSwap, COCO-CF, and Ground-CF. VCF tests object colours, ColorSwap tests colour--object bindings, and COCO-CF tests entity substitutions in natural images. We also construct Ground-CF, a controlled benchmark covering colour, shape, count, and spatial relations. Since our goal is to measure whether a requested source continues to control the answer when the evidence changes, all evaluations use conflicting image--text pairs and counterfactual edits that preserve this conflict. \autoref{tab:verified_benchmarks} summarizes the evaluated subsets and held-out values.

\begin{table}[htbp]
\centering
\scriptsize
\caption{\textbf{Controlled benchmarks and counterfactual interventions.}
``Items'' gives the evaluated conflict examples rather than the size of the original dataset. Ground-CF reports all items, with the eligible image-twin subset in parentheses.}
\label{tab:verified_benchmarks}
\setlength{\tabcolsep}{3pt}
\begin{tabular}{@{}l
>{\raggedright\arraybackslash}p{1.8cm}
>{\raggedright\arraybackslash}p{2.7cm}
>{\raggedright\arraybackslash}p{1.8cm}
c
>{\raggedright\arraybackslash}p{1.7cm}
>{\raggedright\arraybackslash}p{2.6cm}@{}}
\toprule
Benchmark & Queried value & Image edit & Text edit & Cand. & Items ID / Held-out & Held out \\
\midrule
VCF
& object colour
& counterfactual-colour image of the same object
& colour replaced
& 4
& 120 / 140
& 94 objects unseen in training \\

ColorSwap
& colour--object binding
& colours swapped between objects
& colours reassigned
& 4
& 140 / 140
& pink, purple \\

COCO-CF
& entity noun
& counterfactual image
& noun replaced
& 4
& 140 / 140
& nouns unseen in training \\

Ground-CF
& colour, shape, count, spatial relation
& queried value changed; count/spatial scenes re-rendered
& queried value replaced
& 3
& 748 / 800 (707 / 800)
& purple, pink; triangle, star; counts 7--9; \emph{above} \\
\bottomrule
\end{tabular}
\end{table}

\paragraph{Conflict-preserving twins.}
Each example contains an image and text that support different answers. For \textsc{Flip}, we change the requested source while keeping the competing source fixed; for \textsc{Stay}, we change only the competing source. Importantly, the two sources continue to disagree after either intervention. The question, grounding request, and candidate set remain fixed across the original example and its twins. This allows us to measure whether the prediction follows the requested evidence rather than whether an edit merely removes the conflict. For VCF, ColorSwap, and COCO-CF, the candidate set contains the original and edited values from both sources, and the same four candidates are scored in every context.

\paragraph{Ground-CF.}
Ground-CF provides the same intervention structure across four attributes. Colour and shape edits change the queried object's attribute, count edits change the number of queried objects, and spatial edits change the queried relation. Text twins replace only the corresponding textual value. We hold out purple and pink, triangle and star, counts 7--9, and the relation \emph{above} as original image answers. Moreover, the image--spatial and text--count request--attribute combinations are not observed during training, allowing us to separately examine whether the learned request carries across attributes. Ground-CF contains 748 ID and 800 held-out evaluation items; valid image twins are available for 707 and 800 of these items, respectively.

\paragraph{Metrics and denominators.}
Accuracy measures correctness on the original input. \textsc{Flip} requires the model to answer correctly before and after the requested source changes, while \textsc{Stay} requires it to remain correct when only the competing source changes. Since some Ground-CF ID count examples do not have valid image twins, image \textsc{Flip} and text \textsc{Stay} use the eligible image-twin subset; image \textsc{Stay} and text \textsc{Flip} use all examples. Any conditional quantity such as $\textsc{Flip}/A$ therefore recomputes Accuracy on the same eligible items.

\section{Training and Evaluation Details}
\label{app:training}\label{app:templates}

\paragraph{Configurations.}
The controlled experiments use one learned embedding per request ($L{=}1$) for neologisms and prompt tuning. Both requests are learned in the same model. LoRA uses rank $16$ and, in the main comparison, one specialist adapter per requested modality. We additionally evaluate one joint, instruction-conditioned LoRA adapter on Ground-CF. As shown in \autoref{tab:verified_config}, the two lightweight methods use only 4,096--5,120 trainable parameters across both requests, compared with 21.8--59.6 million for the two specialist LoRA adapters.

\begin{table}[htbp]
\centering
\small
\caption{\textbf{Trainable parameters in the controlled experiments.}
LoRA counts include both source-specific adapters; one is active for each request.}
\label{tab:verified_config}
\begin{tabular}{lrrr}
\toprule
Method & Gemma-3-4B & Qwen3.5-4B & Qwen3.5-2B\\
\midrule
Neologism / prompt tuning & 5,120 & 5,120 & 4,096\\
LoRA, two adapters & 59,604,992 & 42,467,328 & 21,823,488\\
LoRA, one joint adapter & 29,802,496 & 21,233,664 & 10,911,744\\
\bottomrule
\end{tabular}
\end{table}

\paragraph{Optimization.}
All learned methods in the main comparison use SFT+CTM and are evaluated at checkpoint 800. \autoref{tab:verified_recipe} summarizes the training settings. For neologisms and prompt tuning, gradients are restricted to the newly added embedding rows. At evaluation, these learned rows are inserted into the original pretrained embedding matrix, so the evaluated model differs from the pretrained model only through the learned control embeddings.

\begin{table}[htbp]
\centering
\footnotesize
\caption{\textbf{Training settings for the controlled experiments.}}
\label{tab:verified_recipe}
\setlength{\tabcolsep}{4pt}
\begin{tabular}{@{}p{2.5cm}p{5.6cm}p{4.9cm}@{}}
\toprule
Setting & Neologism / prompt tuning & LoRA \\
\midrule
Trainable parameters
& added embedding rows
& rank-16 adapters, $\alpha=32$, dropout $.05$ \\

Adapted modules
& ---
& q, k, v, o, gate, up, down projections; vision tower and projector frozen \\

Optimizer
& AdamW, weight decay $.01$
& AdamW, weight decay $.01$ \\

Peak learning rate
& $.05$
& $.0002$ \\

Schedule
& 20 warm-up steps, cosine decay to $10\%$ of peak
& same \\

Steps / checkpoint
& 800 / 800
& 800 / 800 \\

Batch
& 1 example $\times$ 8 accumulation steps; gradient clipping 1
& same \\

Objective
& $\lambda_{\rm SFT}=\lambda_{\rm CTM}=1$, $\gamma=.1$; CTM ramps from 0 at step 100 to 1 at step 150
& same \\

Initialization
& mean embedding of the initialization word's subtokens $+\mathcal N(0,.01^2)$; row norms capped at $2\times$ the mean vocabulary-row norm
& library default ($B=0$) \\
\bottomrule
\end{tabular}
\end{table}

\paragraph{How the request enters the model.}
All methods receive the same image, text, question, candidate answers, and scored continuation. They differ only in how the requested modality is represented. The English baseline states the request directly. Prompt tuning inserts a learned vector at the beginning of the textual input. In contrast, a neologism is a new vocabulary word placed inside a short natural-language request. The specialist LoRA baseline selects the requested modality through the corresponding adapter, while the joint LoRA uses an English control string. \autoref{tab:verified_prompts} gives representative image-request examples; the text request replaces image/\lk{} with caption/\rdw{}.

\begin{table}[htbp]
\centering
\footnotesize
\caption{\textbf{How each modality request is represented.}
Evaluation wordings are not used during training.}
\label{tab:verified_prompts}
\setlength{\tabcolsep}{4pt}
\begin{tabular}{@{}llp{3.3cm}p{5.2cm}@{}}
\toprule
Method & Position & Training example & Evaluation example\\
\midrule
English
& end
& ---
& \emph{According to the image (and only the image), what is the answer?}\\

Prompt tuning
& start
& \texttt{<pt>} + task sentence
& \texttt{<pt>} + held-out task sentence\\

Neologism
& end
& \emph{Give me a \lk{} answer.}
& \emph{Answer in a \lk{} manner.}\\

LoRA, separate
& adapter
& task sentence
& held-out task sentence\\

LoRA, joint
& end
& \emph{Answer using only the image.}
& held-out instruction\\
\bottomrule
\end{tabular}
\end{table}

On VCF, ColorSwap, and COCO-CF, the image is followed by the conflicting text. Ground-CF additionally includes an explicit question, and image/text order alternates across items while remaining fixed across the original example and its twins. MultiVENT-G uses a shared carrier at the end of the prompt for all learned methods, so neologisms and prompt-tuning vectors occupy the same position.

\paragraph{Does placement explain the difference between the lightweight methods?}
\label{app:position}
Since prompt tuning is placed before the evidence while the reference neologism follows it, we test whether position explains their difference. We therefore train an additional Ground-CF neologism at the prompt-tuning position. As shown in \autoref{tab:verified_position}, moving the word to the beginning has little effect on its average visual \textsc{Flip}: $.725$ to $.716$ in ID and $.608$ to $.609$ on held-out values. The word also retains strong text control at either position. On the other hand, prompt tuning remains stronger on Ground-CF visual control. Thus, placement alone does not explain the difference between the two interfaces. This is consistent with MultiVENT-G, where both controls use the same position and neologisms outperform prompt tuning on visual \textsc{Flip} for the two Qwen models.

\begin{table}[htbp]
\centering
\footnotesize
\caption{\textbf{Ground-CF placement control}, averaged across the three models.
Words are evaluated at the position in which they were trained.}
\label{tab:verified_position}
\setlength{\tabcolsep}{4pt}
\begin{tabular}{llcccc}
\toprule
Split & Interface & $F_I$ & $S_I$ & $F_T$ & $S_T$\\
\midrule
ID & Word, end & .725 & .773 & .994 & .996\\
ID & Word, start & .716 & .767 & .988 & .988\\
ID & Prompt tuning, start & .767 & .809 & .979 & .983\\
Held-out & Word, end & .608 & .645 & .995 & .997\\
Held-out & Word, start & .609 & .647 & .981 & .984\\
Held-out & Prompt tuning, start & .630 & .667 & .971 & .977\\
\bottomrule
\end{tabular}
\end{table}

\paragraph{Candidate scoring.}
Since each benchmark provides a fixed set of valid answers, we score the candidates directly rather than relying on open generation. For candidate $y=(y_1,\ldots,y_T)$, we sum its token log probabilities while conditioning on the multimodal input, the grounding request, and the preceding candidate tokens. All candidates use the same answer format, and the highest-scoring candidate is selected. Exact ties are resolved using the fixed candidate order. The reported controlled results are therefore forced-choice likelihood results rather than free-generation accuracy.

\paragraph{Computational resources.}
Each run uses one GPU, primarily an NVIDIA L40S 48\,GB, with A100, H100, or H200 GPUs used for part of the evaluation. On Ground-CF, 800 training steps take approximately 0.7 hours for SFT, 3.5--7.5 hours for SFT+CTM, and 4.9 hours for SFT+CTM+Stay; a full evaluation takes approximately 20--50 minutes.

\section{Neologism Learning and Counterfactual Supervision}
\label{app:neologism}\label{app:proofs}

\paragraph{Neologism learning.}
Let $E\in\mathbb R^{|\mathcal V|\times d}$ denote the input-embedding matrix of the pretrained VLM. For each modality request $s\in\{\text{look},\text{read}\}$, we introduce $L$ new vocabulary entries with learned embeddings $\theta_s\in\mathbb R^{L\times d}$. The new token IDs are used only as inputs and are excluded from output scoring. In the controlled experiments, $L{=}1$, so each request is represented by a single learned embedding. At evaluation, the learned rows are inserted into the original pretrained embedding matrix, so the evaluated model differs from the pretrained model only through these new control embeddings.

\paragraph{Counterfactual Twin-Margin.}
Since standard SFT optimizes each example independently, it does not directly specify how answer preferences should change when the requested evidence changes. CTM instead compares the same answers across an original context $C$ and its requested-source twin $C_s'$. If the requested source supports $v$ before the edit and $v'$ afterwards, we define

$$
\Delta_{\rm old}
=
m_s(v\mid C)-m_s(v\mid C_s'),
\qquad
\Delta_{\rm new}
=
m_s(v'\mid C_s')-m_s(v'\mid C).
$$

We then optimize

$$
\mathcal L_{\rm CTM}
=
-\log\sigma(\gamma\Delta_{\rm old})
-\log\sigma(\gamma\Delta_{\rm new}),
$$

together with the original-example SFT loss. The first margin encourages the old answer to lose log-likelihood after the requested source changes, while the second encourages the new answer to gain log-likelihood.

To show what these two margins imply, define

$$
D(C)=m_s(v\mid C)-m_s(v'\mid C),
$$

the model's preference for the original answer over the replacement. Then

$$
\Delta_{\rm old}+\Delta_{\rm new}
=
D(C)-D(C_s').
$$

Thus, if both margins are positive, the model's relative preference shifts toward $v'$ after the requested-source edit. On the other hand, a positive sum alone does not imply that both individual margins are positive.

\paragraph{What CTM does and does not enforce.}
CTM provides a directional training signal, but it does not guarantee that $v'$ becomes the highest-scoring answer. We therefore still evaluate \textsc{Flip} directly. Moreover, CTM only observes edits to the requested source and does not explicitly train invariance to changes in the competing source. For the Stay ablation, we additionally use

$$
\mathcal L_{\rm Stay}
=
-\log\sigma\!\left(
\gamma[
m_s(v\mid C_u')-m_s(w\mid C_u')
]
\right),
$$

where $C_u'$ changes the competing source to value $w$ while the requested answer remains $v$.

\paragraph{What the objective comparisons test.}
SFT trains only on the original examples, whereas CTM introduces the counterfactual twins together with the cross-context margin. To separate these two effects, paired SFT receives the same original and edited examples but optimizes them independently with cross-entropy. Thus, CTM versus paired SFT tests whether explicitly coupling the two contexts provides additional benefit beyond simply seeing the same counterfactual examples. We report this control for Gemma-3-4B and Qwen3.5-2B and restrict conclusions from this comparison to these two models.


\section{Complete Controlled Results}
\label{app:fulltables}
\label{app:everycell}
\label{app:grids}

\subsection{Full Model--Benchmark Results}
\label{app:full_grid}

Tables~\ref{tab:verified_G3}--\ref{tab:verified_Q2} report every cell used in the controlled main-grid averages. $I$ and $T$ denote image and text requests, while $A$, $F$, and $S$ denote Accuracy, \textsc{Flip}, and \textsc{Stay}. Across the full grid, neologisms improve visual \textsc{Flip} over natural-language instructions in every cell. Prompt tuning and neologisms are closer, and their relative performance depends on the model and benchmark. On the other hand, the substantially larger LoRA adapters provide the strongest average control overall.


\begin{table}[p]
\centering
\scriptsize
\caption{\textbf{Complete Gemma-3-4B results.}
Ground-CF visual \textsc{Flip} and text \textsc{Stay} use the eligible image-twin subset; all other metrics use the full evaluation set. Joint LoRA is evaluated only on Ground-CF.}
\label{tab:verified_G3}
\setlength{\tabcolsep}{3pt}
\resizebox{\linewidth}{!}{%
\begin{tabular}{lllcccccc}
\toprule
Split & Dataset & Method & $A_I$ & $F_I$ & $S_I$ & $A_T$ & $F_T$ & $S_T$ \\
\midrule
ID & VCF & No request & 0.167 & 0.017 & 0.100 & 0.833 & 0.800 & 0.683 \\
ID & VCF & Instruction & 0.608 & 0.367 & 0.517 & 0.983 & 0.967 & 0.967 \\
ID & VCF & Prompt tuning & 0.817 & 0.700 & 0.700 & 1.000 & 1.000 & 1.000 \\
ID & VCF & Neologism & 0.933 & 0.867 & 0.900 & 1.000 & 1.000 & 1.000 \\
ID & VCF & LoRA (separate) & 0.892 & 0.783 & 0.842 & 1.000 & 1.000 & 1.000 \\

ID & ColorSwap & No request & 0.407 & 0.129 & 0.250 & 0.571 & 0.471 & 0.493 \\
ID & ColorSwap & Instruction & 0.536 & 0.336 & 0.421 & 0.771 & 0.679 & 0.729 \\
ID & ColorSwap & Prompt tuning & 0.836 & 0.693 & 0.779 & 0.993 & 0.979 & 0.993 \\
ID & ColorSwap & Neologism & 0.771 & 0.586 & 0.707 & 0.786 & 0.693 & 0.757 \\
ID & ColorSwap & LoRA (separate) & 0.921 & 0.400 & 0.857 & 0.993 & 0.993 & 0.993 \\

ID & COCO-CF & No request & 0.214 & 0.057 & 0.157 & 0.579 & 0.371 & 0.500 \\
ID & COCO-CF & Instruction & 0.336 & 0.150 & 0.257 & 0.793 & 0.729 & 0.729 \\
ID & COCO-CF & Prompt tuning & 0.050 & 0.014 & 0.014 & 0.993 & 0.993 & 0.993 \\
ID & COCO-CF & Neologism & 0.607 & 0.379 & 0.479 & 1.000 & 1.000 & 1.000 \\
ID & COCO-CF & LoRA (separate) & 0.943 & 0.886 & 0.929 & 1.000 & 0.993 & 1.000 \\

ID & Ground-CF & No request & 0.457 & 0.300 & 0.349 & 0.477 & 0.320 & 0.376 \\
ID & Ground-CF & Instruction & 0.480 & 0.315 & 0.384 & 0.849 & 0.761 & 0.785 \\
ID & Ground-CF & Prompt tuning & 0.766 & 0.655 & 0.741 & 0.964 & 0.941 & 0.949 \\
ID & Ground-CF & Neologism & 0.759 & 0.632 & 0.693 & 0.996 & 0.983 & 0.987 \\
ID & Ground-CF & LoRA (separate) & 0.862 & 0.755 & 0.840 & 1.000 & 1.000 & 1.000 \\
ID & Ground-CF & LoRA (joint) & 0.825 & 0.741 & 0.802 & 0.837 & 0.807 & 0.833 \\

\midrule

Held-out & VCF & No request & 0.186 & 0.071 & 0.129 & 0.814 & 0.736 & 0.700 \\
Held-out & VCF & Instruction & 0.529 & 0.271 & 0.457 & 1.000 & 0.986 & 1.000 \\
Held-out & VCF & Prompt tuning & 0.814 & 0.700 & 0.714 & 1.000 & 1.000 & 1.000 \\
Held-out & VCF & Neologism & 0.943 & 0.886 & 0.907 & 1.000 & 1.000 & 1.000 \\
Held-out & VCF & LoRA (separate) & 0.950 & 0.900 & 0.907 & 1.000 & 1.000 & 1.000 \\

Held-out & ColorSwap & No request & 0.557 & 0.300 & 0.379 & 0.371 & 0.286 & 0.286 \\
Held-out & ColorSwap & Instruction & 0.757 & 0.479 & 0.664 & 0.607 & 0.550 & 0.571 \\
Held-out & ColorSwap & Prompt tuning & 0.900 & 0.779 & 0.864 & 0.979 & 0.979 & 0.979 \\
Held-out & ColorSwap & Neologism & 0.886 & 0.700 & 0.821 & 0.600 & 0.507 & 0.564 \\
Held-out & ColorSwap & LoRA (separate) & 0.936 & 0.386 & 0.886 & 0.993 & 0.986 & 0.993 \\

Held-out & COCO-CF & No request & 0.350 & 0.193 & 0.221 & 0.514 & 0.429 & 0.436 \\
Held-out & COCO-CF & Instruction & 0.436 & 0.207 & 0.307 & 0.779 & 0.664 & 0.693 \\
Held-out & COCO-CF & Prompt tuning & 0.079 & 0.014 & 0.036 & 0.993 & 0.979 & 0.986 \\
Held-out & COCO-CF & Neologism & 0.586 & 0.400 & 0.443 & 0.979 & 0.979 & 0.979 \\
Held-out & COCO-CF & LoRA (separate) & 0.871 & 0.757 & 0.857 & 1.000 & 1.000 & 1.000 \\

Held-out & Ground-CF & No request & 0.531 & 0.415 & 0.414 & 0.351 & 0.264 & 0.305 \\
Held-out & Ground-CF & Instruction & 0.537 & 0.440 & 0.449 & 0.770 & 0.695 & 0.726 \\
Held-out & Ground-CF & Prompt tuning & 0.664 & 0.610 & 0.634 & 0.940 & 0.914 & 0.930 \\
Held-out & Ground-CF & Neologism & 0.650 & 0.604 & 0.591 & 0.996 & 0.995 & 0.996 \\
Held-out & Ground-CF & LoRA (separate) & 0.654 & 0.584 & 0.635 & 1.000 & 1.000 & 1.000 \\
Held-out & Ground-CF & LoRA (joint) & 0.581 & 0.560 & 0.565 & 0.853 & 0.809 & 0.849 \\
\bottomrule
\end{tabular}
}
\end{table}


\begin{table}[p]
\centering
\scriptsize
\caption{\textbf{Complete Qwen3.5-4B results.}
Conventions follow \autoref{tab:verified_G3}.}
\label{tab:verified_Q4}
\setlength{\tabcolsep}{3pt}
\resizebox{\linewidth}{!}{%
\begin{tabular}{lllcccccc}
\toprule
Split & Dataset & Method & $A_I$ & $F_I$ & $S_I$ & $A_T$ & $F_T$ & $S_T$ \\
\midrule
ID & VCF & No request & 0.267 & 0.117 & 0.150 & 0.733 & 0.675 & 0.583 \\
ID & VCF & Instruction & 0.700 & 0.550 & 0.625 & 1.000 & 1.000 & 1.000 \\
ID & VCF & Prompt tuning & 0.850 & 0.733 & 0.758 & 1.000 & 1.000 & 1.000 \\
ID & VCF & Neologism & 0.908 & 0.817 & 0.875 & 1.000 & 1.000 & 1.000 \\
ID & VCF & LoRA (separate) & 0.883 & 0.767 & 0.825 & 1.000 & 1.000 & 1.000 \\

ID & ColorSwap & No request & 0.529 & 0.300 & 0.357 & 0.429 & 0.257 & 0.307 \\
ID & ColorSwap & Instruction & 0.636 & 0.464 & 0.493 & 0.957 & 0.907 & 0.936 \\
ID & ColorSwap & Prompt tuning & 0.886 & 0.721 & 0.850 & 0.979 & 0.964 & 0.979 \\
ID & ColorSwap & Neologism & 0.914 & 0.786 & 0.850 & 0.943 & 0.893 & 0.936 \\
ID & ColorSwap & LoRA (separate) & 0.893 & 0.807 & 0.850 & 1.000 & 1.000 & 1.000 \\

ID & COCO-CF & No request & 0.271 & 0.164 & 0.179 & 0.671 & 0.471 & 0.600 \\
ID & COCO-CF & Instruction & 0.336 & 0.136 & 0.229 & 0.957 & 0.943 & 0.943 \\
ID & COCO-CF & Prompt tuning & 0.464 & 0.307 & 0.386 & 1.000 & 0.993 & 0.993 \\
ID & COCO-CF & Neologism & 0.636 & 0.471 & 0.564 & 1.000 & 1.000 & 1.000 \\
ID & COCO-CF & LoRA (separate) & 0.950 & 0.900 & 0.929 & 1.000 & 1.000 & 1.000 \\

ID & Ground-CF & No request & 0.711 & 0.586 & 0.622 & 0.217 & 0.088 & 0.117 \\
ID & Ground-CF & Instruction & 0.750 & 0.628 & 0.672 & 0.949 & 0.866 & 0.917 \\
ID & Ground-CF & Prompt tuning & 0.888 & 0.846 & 0.873 & 1.000 & 0.997 & 1.000 \\
ID & Ground-CF & Neologism & 0.882 & 0.789 & 0.846 & 1.000 & 1.000 & 1.000 \\
ID & Ground-CF & LoRA (separate) & 0.989 & 0.973 & 0.987 & 1.000 & 1.000 & 1.000 \\
ID & Ground-CF & LoRA (joint) & 0.925 & 0.870 & 0.910 & 1.000 & 0.996 & 0.999 \\

\midrule

Held-out & VCF & No request & 0.179 & 0.071 & 0.114 & 0.821 & 0.700 & 0.714 \\
Held-out & VCF & Instruction & 0.664 & 0.471 & 0.543 & 1.000 & 1.000 & 1.000 \\
Held-out & VCF & Prompt tuning & 0.864 & 0.757 & 0.757 & 1.000 & 1.000 & 1.000 \\
Held-out & VCF & Neologism & 0.914 & 0.843 & 0.829 & 1.000 & 1.000 & 1.000 \\
Held-out & VCF & LoRA (separate) & 0.921 & 0.843 & 0.850 & 1.000 & 1.000 & 1.000 \\

Held-out & ColorSwap & No request & 0.507 & 0.379 & 0.386 & 0.450 & 0.343 & 0.257 \\
Held-out & ColorSwap & Instruction & 0.721 & 0.643 & 0.629 & 0.986 & 0.979 & 0.986 \\
Held-out & ColorSwap & Prompt tuning & 0.929 & 0.800 & 0.893 & 0.979 & 0.979 & 0.979 \\
Held-out & ColorSwap & Neologism & 0.886 & 0.800 & 0.857 & 0.929 & 0.914 & 0.921 \\
Held-out & ColorSwap & LoRA (separate) & 0.850 & 0.764 & 0.707 & 0.986 & 0.986 & 0.986 \\

Held-out & COCO-CF & No request & 0.350 & 0.136 & 0.250 & 0.600 & 0.550 & 0.443 \\
Held-out & COCO-CF & Instruction & 0.421 & 0.193 & 0.350 & 0.943 & 0.893 & 0.921 \\
Held-out & COCO-CF & Prompt tuning & 0.536 & 0.329 & 0.464 & 0.993 & 0.986 & 0.993 \\
Held-out & COCO-CF & Neologism & 0.686 & 0.471 & 0.636 & 0.979 & 0.979 & 0.971 \\
Held-out & COCO-CF & LoRA (separate) & 0.879 & 0.764 & 0.857 & 1.000 & 1.000 & 1.000 \\

Held-out & Ground-CF & No request & 0.581 & 0.554 & 0.535 & 0.242 & 0.142 & 0.176 \\
Held-out & Ground-CF & Instruction & 0.601 & 0.571 & 0.552 & 0.955 & 0.915 & 0.919 \\
Held-out & Ground-CF & Prompt tuning & 0.627 & 0.585 & 0.601 & 1.000 & 1.000 & 1.000 \\
Held-out & Ground-CF & Neologism & 0.660 & 0.600 & 0.621 & 1.000 & 0.994 & 1.000 \\
Held-out & Ground-CF & LoRA (separate) & 0.786 & 0.718 & 0.767 & 1.000 & 1.000 & 1.000 \\
Held-out & Ground-CF & LoRA (joint) & 0.735 & 0.680 & 0.716 & 0.999 & 0.999 & 0.999 \\
\bottomrule
\end{tabular}
}
\end{table}


\begin{table}[p]
\centering
\scriptsize
\caption{\textbf{Complete Qwen3.5-2B results.}
Conventions follow \autoref{tab:verified_G3}.}
\label{tab:verified_Q2}
\setlength{\tabcolsep}{3pt}
\resizebox{\linewidth}{!}{%
\begin{tabular}{lllcccccc}
\toprule
Split & Dataset & Method & $A_I$ & $F_I$ & $S_I$ & $A_T$ & $F_T$ & $S_T$ \\
\midrule
ID & VCF & No request & 0.417 & 0.150 & 0.317 & 0.583 & 0.483 & 0.317 \\
ID & VCF & Instruction & 0.633 & 0.367 & 0.525 & 0.833 & 0.758 & 0.683 \\
ID & VCF & Prompt tuning & 0.933 & 0.867 & 0.900 & 1.000 & 1.000 & 1.000 \\
ID & VCF & Neologism & 0.875 & 0.767 & 0.792 & 1.000 & 1.000 & 1.000 \\
ID & VCF & LoRA (separate) & 0.850 & 0.733 & 0.742 & 1.000 & 1.000 & 1.000 \\

ID & ColorSwap & No request & 0.557 & 0.143 & 0.471 & 0.400 & 0.279 & 0.357 \\
ID & ColorSwap & Instruction & 0.757 & 0.479 & 0.686 & 0.493 & 0.371 & 0.464 \\
ID & ColorSwap & Prompt tuning & 0.857 & 0.700 & 0.750 & 0.814 & 0.743 & 0.800 \\
ID & ColorSwap & Neologism & 0.914 & 0.793 & 0.843 & 0.550 & 0.443 & 0.479 \\
ID & ColorSwap & LoRA (separate) & 0.836 & 0.771 & 0.786 & 1.000 & 1.000 & 1.000 \\

ID & COCO-CF & No request & 0.164 & 0.079 & 0.086 & 0.786 & 0.679 & 0.721 \\
ID & COCO-CF & Instruction & 0.236 & 0.114 & 0.129 & 0.829 & 0.743 & 0.764 \\
ID & COCO-CF & Prompt tuning & 0.693 & 0.521 & 0.629 & 1.000 & 1.000 & 1.000 \\
ID & COCO-CF & Neologism & 0.579 & 0.329 & 0.464 & 1.000 & 0.993 & 1.000 \\
ID & COCO-CF & LoRA (separate) & 0.957 & 0.914 & 0.893 & 1.000 & 1.000 & 0.993 \\

ID & Ground-CF & No request & 0.620 & 0.458 & 0.488 & 0.271 & 0.164 & 0.167 \\
ID & Ground-CF & Instruction & 0.658 & 0.496 & 0.527 & 0.765 & 0.686 & 0.666 \\
ID & Ground-CF & Prompt tuning & 0.861 & 0.801 & 0.814 & 1.000 & 1.000 & 1.000 \\
ID & Ground-CF & Neologism & 0.826 & 0.754 & 0.779 & 1.000 & 1.000 & 1.000 \\
ID & Ground-CF & LoRA (separate) & 0.957 & 0.917 & 0.953 & 1.000 & 1.000 & 1.000 \\
ID & Ground-CF & LoRA (joint) & 0.917 & 0.853 & 0.910 & 0.814 & 0.793 & 0.829 \\

\midrule

Held-out & VCF & No request & 0.314 & 0.157 & 0.271 & 0.686 & 0.564 & 0.529 \\
Held-out & VCF & Instruction & 0.464 & 0.243 & 0.414 & 0.864 & 0.786 & 0.757 \\
Held-out & VCF & Prompt tuning & 0.964 & 0.929 & 0.943 & 1.000 & 1.000 & 1.000 \\
Held-out & VCF & Neologism & 0.836 & 0.686 & 0.779 & 1.000 & 1.000 & 1.000 \\
Held-out & VCF & LoRA (separate) & 0.857 & 0.714 & 0.757 & 1.000 & 1.000 & 1.000 \\

Held-out & ColorSwap & No request & 0.721 & 0.279 & 0.629 & 0.264 & 0.157 & 0.207 \\
Held-out & ColorSwap & Instruction & 0.871 & 0.657 & 0.793 & 0.407 & 0.286 & 0.379 \\
Held-out & ColorSwap & Prompt tuning & 0.886 & 0.757 & 0.829 & 0.886 & 0.821 & 0.879 \\
Held-out & ColorSwap & Neologism & 0.929 & 0.843 & 0.900 & 0.586 & 0.479 & 0.536 \\
Held-out & ColorSwap & LoRA (separate) & 0.664 & 0.593 & 0.593 & 0.993 & 0.986 & 0.993 \\

Held-out & COCO-CF & No request & 0.257 & 0.114 & 0.143 & 0.721 & 0.664 & 0.607 \\
Held-out & COCO-CF & Instruction & 0.329 & 0.200 & 0.221 & 0.779 & 0.743 & 0.693 \\
Held-out & COCO-CF & Prompt tuning & 0.764 & 0.600 & 0.693 & 1.000 & 0.993 & 1.000 \\
Held-out & COCO-CF & Neologism & 0.621 & 0.429 & 0.500 & 0.986 & 0.979 & 0.979 \\
Held-out & COCO-CF & LoRA (separate) & 0.864 & 0.743 & 0.836 & 1.000 & 1.000 & 1.000 \\

Held-out & Ground-CF & No request & 0.497 & 0.344 & 0.414 & 0.324 & 0.169 & 0.200 \\
Held-out & Ground-CF & Instruction & 0.544 & 0.434 & 0.454 & 0.812 & 0.651 & 0.762 \\
Held-out & Ground-CF & Prompt tuning & 0.786 & 0.696 & 0.767 & 1.000 & 1.000 & 1.000 \\
Held-out & Ground-CF & Neologism & 0.755 & 0.619 & 0.723 & 0.998 & 0.995 & 0.995 \\
Held-out & Ground-CF & LoRA (separate) & 0.759 & 0.709 & 0.749 & 1.000 & 1.000 & 1.000 \\
Held-out & Ground-CF & LoRA (joint) & 0.708 & 0.671 & 0.698 & 0.800 & 0.790 & 0.794 \\
\bottomrule
\end{tabular}
}
\end{table}


\subsection{Ground-CF Results by Attribute}
\label{app:attributes}

Since Ground-CF covers four different attributes, we additionally separate colour, shape, count, and spatial-relation questions. The largest differences appear for visual grounding. Neologisms remain particularly strong on colour and shape, while count and spatial relations are harder, especially on held-out values. Text control, on the other hand, remains substantially stronger across attributes.

This breakdown also explains why the cross-benchmark held-out average can remain stable even though Ground-CF itself becomes harder. For neologisms, visual \textsc{Flip} decreases from $.725$ to $.608$ on Ground-CF when moving from ID to held-out evaluation, with much of the difficulty concentrated in count and spatial-relation questions. Moreover, the image--spatial and text--count request--attribute combinations are not observed during training, allowing us to test whether the learned modality request carries across attributes.

\subsection{Control Conditional on an Initially Correct Answer}
\label{app:conditional}

Since \textsc{Flip} and \textsc{Stay} require the original prediction to be correct, we also examine control conditional on this initial correctness. \autoref{tab:conditional_instruction} reports $F/A$ and $S/A$ for the natural-language instruction, with Accuracy computed over the same eligible examples. For visual requests, between $16\%$ and $60\%$ of initially correct answers still fail to follow the image edit. On the other hand, text requests are generally more reliable. Thus, the gap between correctness and control remains even after conditioning on an initially correct prediction.

\begin{table}[htbp]
\centering
\footnotesize
\caption{\textbf{Conditional control for natural-language instructions in ID evaluation.}
$F/A$ and $S/A$ give the fraction of initially correct predictions that remain correct after the corresponding intervention.}
\label{tab:conditional_instruction}
\setlength{\tabcolsep}{4pt}
\begin{tabular}{llcccc}
\toprule
Model & Benchmark & $F_I/A_I$ & $S_I/A_I$ & $F_T/A_T$ & $S_T/A_T$ \\
\midrule
Gemma-3-4B & VCF & .603 & .849 & .983 & .983 \\
 & ColorSwap & .627 & .787 & .880 & .944 \\
 & COCO-CF & .447 & .766 & .919 & .919 \\
 & Ground-CF & .639 & .799 & .896 & .934 \\
Qwen3.5-4B & VCF & .786 & .893 & 1.000 & 1.000 \\
 & ColorSwap & .730 & .775 & .948 & .978 \\
 & COCO-CF & .404 & .681 & .985 & .985 \\
 & Ground-CF & .843 & .897 & .913 & .969 \\
Qwen3.5-2B & VCF & .579 & .829 & .910 & .820 \\
 & ColorSwap & .632 & .906 & .754 & .942 \\
 & COCO-CF & .485 & .545 & .897 & .922 \\
 & Ground-CF & .778 & .801 & .897 & .885 \\
\bottomrule
\end{tabular}
\end{table}


\subsection{Objective Comparison}
\label{app:objective}

We compare SFT, paired SFT, SFT+CTM, and SFT+CTM+Stay on Ground-CF in \autoref{tab:verified_objectives}. CTM provides its clearest gains where standard supervision is weaker. In particular, on the two models with paired-SFT controls, CTM improves held-out visual control beyond exposure to the same counterfactual examples alone: by $+.065/+.059$ \textsc{Flip}/\textsc{Stay} for Gemma-3-4B and $+.031/+.115$ for Qwen3.5-2B. On Gemma-3-4B, the visual \textsc{Flip} gain also appears in ID evaluation.

On the other hand, adding explicit \textsc{Stay} supervision does not consistently improve control. Although it helps some visual settings, it can substantially reduce text control. For example, Qwen3.5-4B text \textsc{Flip} falls from $1.000$ to $.512$ in ID evaluation and from $.994$ to $.610$ on held-out values. We therefore use CTM without the additional Stay term as the reference objective.

\begin{table}[htbp]
\centering
\scriptsize
\caption{\textbf{Ground-CF objective comparison} for $L{=}1$ neologisms at checkpoint 800. G3, Q4, and Q2 denote Gemma-3-4B, Qwen3.5-4B, and Qwen3.5-2B. Paired SFT is available for Gemma-3-4B and Qwen3.5-2B.}
\label{tab:verified_objectives}
\setlength{\tabcolsep}{3pt}
\resizebox{.9\linewidth}{!}{%
\begin{tabular}{lllcccccc}
\toprule
Split & Model & Objective & $A_I$ & $F_I$ & $S_I$ & $A_T$ & $F_T$ & $S_T$ \\
\midrule
ID & G3 & SFT & 0.783 & 0.605 & 0.738 & 0.959 & 0.914 & 0.936 \\
ID & G3 & Paired SFT & 0.734 & 0.571 & 0.686 & 0.971 & 0.947 & 0.941 \\
ID & G3 & SFT+CTM & 0.759 & 0.632 & 0.693 & 0.996 & 0.983 & 0.987 \\
ID & G3 & SFT+CTM+Stay & 0.779 & 0.621 & 0.723 & 0.995 & 0.989 & 0.989 \\

ID & Q4 & SFT & 0.755 & 0.666 & 0.687 & 1.000 & 1.000 & 1.000 \\
ID & Q4 & Paired SFT & -- & -- & -- & -- & -- & -- \\
ID & Q4 & SFT+CTM & 0.882 & 0.789 & 0.846 & 1.000 & 1.000 & 1.000 \\
ID & Q4 & SFT+CTM+Stay & 0.896 & 0.842 & 0.861 & 0.650 & 0.512 & 0.547 \\

ID & Q2 & SFT & 0.852 & 0.778 & 0.793 & 1.000 & 0.999 & 1.000 \\
ID & Q2 & Paired SFT & 0.850 & 0.777 & 0.782 & 0.995 & 0.988 & 0.992 \\
ID & Q2 & SFT+CTM & 0.826 & 0.754 & 0.779 & 1.000 & 1.000 & 1.000 \\
ID & Q2 & SFT+CTM+Stay & 0.869 & 0.786 & 0.805 & 1.000 & 1.000 & 1.000 \\

\midrule

Held-out & G3 & SFT & 0.584 & 0.541 & 0.555 & 0.932 & 0.894 & 0.921 \\
Held-out & G3 & Paired SFT & 0.566 & 0.539 & 0.532 & 0.976 & 0.956 & 0.970 \\
Held-out & G3 & SFT+CTM & 0.650 & 0.604 & 0.591 & 0.996 & 0.995 & 0.996 \\
Held-out & G3 & SFT+CTM+Stay & 0.641 & 0.569 & 0.613 & 0.998 & 0.994 & 0.995 \\

Held-out & Q4 & SFT & 0.585 & 0.542 & 0.544 & 1.000 & 0.999 & 1.000 \\
Held-out & Q4 & Paired SFT & -- & -- & -- & -- & -- & -- \\
Held-out & Q4 & SFT+CTM & 0.660 & 0.600 & 0.621 & 1.000 & 0.994 & 1.000 \\
Held-out & Q4 & SFT+CTM+Stay & 0.704 & 0.677 & 0.686 & 0.710 & 0.610 & 0.634 \\

Held-out & Q2 & SFT & 0.659 & 0.645 & 0.625 & 1.000 & 1.000 & 1.000 \\
Held-out & Q2 & Paired SFT & 0.649 & 0.588 & 0.608 & 0.988 & 0.980 & 0.981 \\
Held-out & Q2 & SFT+CTM & 0.755 & 0.619 & 0.723 & 0.998 & 0.995 & 0.995 \\
Held-out & Q2 & SFT+CTM+Stay & 0.690 & 0.596 & 0.662 & 1.000 & 1.000 & 1.000 \\
\bottomrule
\end{tabular}
}
\end{table}

\subsection{Generalisation across prompt formats}
\label{app:prompt-transfer}

Since a modality request can be expressed in different ways,
we additionally examine sensitivity to question wording, request
wording, and image--caption order. These checks complement the
held-out-value evaluation: changing the presentation of a request
is different from changing the content the model must identify.
The supplementary comparisons include $L=5$ input-level controls
and joint rank-16 LoRA, and are reported separately from the
main-grid averages.

\subsubsection{Held-out request carriers and question wording}

We change the question wording, the request carrier, or both.
The carrier is the sentence expressing the modality request,
rather than the question about the image or caption.
The evidence, requested modality, and candidate answers remain
unchanged, and each wording change is applied consistently to
the original example and its counterfactual twins.
For joint LoRA, we rephrase the natural-language request while
keeping the same adapter active.

\autoref{tab:lora-wording} shows little sensitivity to these
wording changes in the completed joint-LoRA comparisons.
Across the reported configurations, the largest absolute change
in image \textsc{Flip} from the corresponding reference is
$.024$, while caption \textsc{Flip} remains at or near ceiling.
Thus, the larger held-out-value degradation reported in the
main comparison should not be interpreted as poor generalisation
to prompt wording. These results also do not establish a
wording-transfer advantage for neologisms over joint LoRA.

\begin{table}[t]
\centering
\small
\setlength{\tabcolsep}{4pt}
\begin{tabular}{@{}llrrrr@{}}
\toprule
Model & Objective
& Reference
& Question
& Carrier
& Both \\
\midrule
Gemma-3-4B & Paired SFT
& .629 & .635 & .642 & .638 \\
Gemma-3-4B & +CTM
& .641 & .617 & .646 & .622 \\
Qwen-4B & Paired SFT
& .921 & .914 & .925 & .914 \\
\bottomrule
\end{tabular}
\caption{
\textbf{Joint LoRA retains control under the tested wording changes.}
Image \textsc{Flip} in the supplementary Ground-CF wording
evaluation. Question and Carrier change the question wording
and natural-language modality request, respectively;
Both applies the two changes together.
The trained adapter remains fixed and active in every condition.
Values are point estimates, so small differences are not
interpreted as statistical equivalence.
}
\label{tab:lora-wording}
\end{table}

\subsubsection{Sensitivity to image--caption order}
\label{app:order}

Ground-CF includes both image-first and caption-first inputs
during training and evaluation. We therefore separate the
results by source order to examine differences that the balanced
average may hide. The order remains fixed across each original
example and its counterfactual twins.
This is an order-stratified analysis, rather than a paired
experiment that reverses the source order of every item.

\autoref{tab:order_split} reports caption \textsc{Flip} for
Gemma-3-4B. The frozen model shows substantial differences
between the two orderings, and paired SFT does not fully remove
these differences for neologisms.
At $L=5$, the ID caption-control gap is $.060$ under paired
SFT, compared with $.000$ at the reported precision under
CTM and $.011$ under CTM+Stay.
The observed gaps are also smaller on held-out values.
This reduction accompanies an improvement in caption control
overall, rather than establishing a separate mechanism for
order invariance.

On the other hand, prompt tuning already shows small
caption-control gaps under paired SFT, and joint LoRA reaches
$1.000$ caption \textsc{Flip} in both orders.
Image-request gaps are at most $.05$ for the learned controls
in this comparison.
Reduced order sensitivity is therefore not specific to
neologisms. Moreover, the $L=1$ neologism retains larger
caption-control gaps under CTM+Stay, so the $L=5$ result
should not be generalized to every control length.

Overall, counterfactual supervision reduces the observed
dependence on source order for the tested five-token neologisms.
This supports the objective comparison by showing that its
benefits can extend across the two evaluated prompt layouts.
It does not establish that neologisms are more robust than
prompt tuning or LoRA, or that a learned request can be moved
to arbitrary positions without loss.

\begin{table}[t]
\centering
\small
\setlength{\tabcolsep}{4pt}
\begin{tabular}{@{}lrrrr@{}}
\toprule
& \multicolumn{2}{c}{ID}
& \multicolumn{2}{c}{Held-out values} \\
\cmidrule(lr){2-3}
\cmidrule(l){4-5}
Configuration
& Caption first & Image first
& Caption first & Image first \\
\midrule
Frozen, no request
& .561 & .495 & .527 & .350 \\
Neologism, $L=5$, paired SFT
& .953 & .893 & .927 & .893 \\
Neologism, $L=5$, +CTM
& .977 & .977 & .967 & .983 \\
Neologism, $L=5$, +CTM+Stay
& .993 & .982 & .989 & .984 \\
Neologism, $L=1$, +CTM+Stay
& .949 & .855 & .945 & .907 \\
Prompt tuning, $L=5$, paired SFT
& .863 & .849 & .880 & .863 \\
Joint LoRA
& 1.000 & 1.000 & 1.000 & 1.000 \\
\bottomrule
\end{tabular}
\caption{
\textbf{The observed source-order gap depends on the control
configuration and objective.}
Caption \textsc{Flip} on Ground-CF for Gemma-3-4B,
separated by image--caption order.
The $L=5$ neologism rows and the paired-SFT prompt-tuning
row average three training seeds.
Values are rounded point estimates; identical scores do not
establish statistical equivalence.
Both source orders occur during training, so this comparison
measures differences across familiar layouts rather than
transfer to unseen positions.
}
\label{tab:order_split}
\end{table}


\subsection{Uncertainty and Aggregation}
\label{app:stats}

The main averages give equal weight to each model--benchmark cell. We use paired exact McNemar tests when two methods are evaluated on the same examples, and item-level bootstrap intervals for the aggregate change from ID to held-out evaluation. \autoref{tab:verified_uncertainty} shows that neologism and prompt-tuning visual control change little on average, whereas LoRA loses more of its visual advantage. These are aggregate results and should therefore not be interpreted as uniform behavior in every benchmark.

\begin{table}[p]
\centering
\scriptsize
\caption{\textbf{Change from ID to held-out evaluation.}
We report macro-average changes with item-bootstrap $95\%$ intervals.}
\label{tab:verified_uncertainty}
\setlength{\tabcolsep}{3pt}
\resizebox{.95\linewidth}{!}{%
\begin{tabular}{lllcc}
\toprule
Request & Metric & Method & Held-out $-$ ID & 95\% interval \\
\midrule
look & flip & Instruction & +0.034 & [+0.007, +0.062] \\
look & flip & Prompt tuning & -0.000 & [-0.026, +0.025] \\
look & flip & Neologism & -0.007 & [-0.033, +0.020] \\
look & flip & LoRA & -0.094 & [-0.118, -0.069] \\

look & stay & Instruction & +0.031 & [+0.003, +0.060] \\
look & stay & Prompt tuning & +0.000 & [-0.025, +0.024] \\
look & stay & Neologism & -0.015 & [-0.042, +0.010] \\
look & stay & LoRA & -0.086 & [-0.108, -0.064] \\

read & flip & Instruction & -0.022 & [-0.043, -0.000] \\
read & flip & Prompt tuning & +0.003 & [-0.007, +0.013] \\
read & flip & Neologism & -0.015 & [-0.031, -0.000] \\
read & flip & LoRA & -0.002 & [-0.006, +0.001] \\

read & stay & Instruction & -0.015 & [-0.036, +0.007] \\
read & stay & Prompt tuning & +0.003 & [-0.005, +0.012] \\
read & stay & Neologism & -0.018 & [-0.033, -0.003] \\
read & stay & LoRA & -0.001 & [-0.004, +0.001] \\
\bottomrule
\end{tabular}
}
\end{table}

\section{Real-World News-Frame Evaluation}
\label{app:mvg}\label{app:applications}

MultiVENT-G uses separate checkpoints trained directly in this setting; the controlled-benchmark checkpoints are not transferred here. Neologisms and prompt tuning use $L{=}5$, while LoRA uses one joint rank-$16$ adapter. We evaluate 173 visual-swap pairs from 22 videos and 300 description-swap pairs. None of the evaluation videos appears in the training set used for the learned methods.

For each pair, we score the same two candidates before and after the intervention. Since \textsc{Flip} requires both predictions to be correct, two independent uniform guesses give an expected score of $.25$. This provides an independent-guessing reference rather than a general chance threshold.

\paragraph{Evidence crops.}
Since MultiVENT-G provides human-annotated visual evidence, we also evaluate a crop centered on the annotated evidence region. The crop is enlarged by $25\%$ to retain surrounding visual context, and no crop-specific training is used. This allows us to test what happens when the relevant visual evidence is easier to access; it does not test whether the model can find that region itself.

\autoref{tab:verified_mvg} reports Accuracy, \textsc{Flip}, and answer stability $U$. Accuracy averages correctness over both versions of the edited-source pair. $U$ measures whether the answer remains unchanged when the competing source changes; unlike \textsc{Stay}, it is not conditioned on correctness. At full frame, visual Accuracy remains between $.50$ and $.66$ across all methods and models, while \textsc{Flip} is often substantially lower. Thus, the main difficulty is not only answering an individual frame correctly, but reliably changing the answer when the relevant visual evidence changes.

\begin{table}[p]
\centering
\scriptsize
\caption{\textbf{Complete MultiVENT-G results.}
We report visual and text Accuracy ($A$), \textsc{Flip} ($F$), and answer stability ($U$) for the full frame and evidence crop. $U$ is answer agreement rather than correctness-conditioned \textsc{Stay}. Visual pairs have $n{=}173$ and text pairs $n{=}300$.}
\label{tab:verified_mvg}
\setlength{\tabcolsep}{3pt}
\resizebox{\linewidth}{!}{%
\begin{tabular}{llcccccc|cccccc}
\toprule
&
&
\multicolumn{6}{c}{Full frame}
&
\multicolumn{6}{c}{Evidence crop}\\
\cmidrule(lr){3-8}
\cmidrule(lr){9-14}
Model & Method
& $A_I$ & $F_I$ & $U_I$ & $A_T$ & $F_T$ & $U_T$
& $A_I$ & $F_I$ & $U_I$ & $A_T$ & $F_T$ & $U_T$ \\
\midrule
Gemma-3-4B & No request & 0.569 & 0.214 & 0.343 & 0.824 & 0.657 & 0.728 & 0.601 & 0.277 & 0.353 & 0.824 & 0.657 & 0.642 \\
Gemma-3-4B & Instruction & 0.573 & 0.191 & 0.330 & 0.860 & 0.727 & 0.751 & 0.593 & 0.266 & 0.327 & 0.862 & 0.727 & 0.613 \\
Gemma-3-4B & Prompt tuning & 0.575 & 0.173 & 0.427 & 0.935 & 0.870 & 0.873 & 0.615 & 0.249 & 0.427 & 0.935 & 0.870 & 0.902 \\
Gemma-3-4B & Neologism & 0.537 & 0.104 & 0.507 & 0.931 & 0.863 & 0.919 & 0.602 & 0.220 & 0.483 & 0.939 & 0.877 & 0.873 \\
Gemma-3-4B & LoRA & 0.575 & 0.191 & 0.430 & 0.992 & 0.983 & 0.827 & 0.702 & 0.416 & 0.450 & 0.994 & 0.987 & 0.769 \\
Qwen3.5-4B & No request & 0.561 & 0.145 & 0.307 & 0.847 & 0.693 & 0.844 & 0.593 & 0.202 & 0.280 & 0.860 & 0.720 & 0.792 \\
Qwen3.5-4B & Instruction & 0.578 & 0.162 & 0.303 & 0.910 & 0.820 & 0.855 & 0.679 & 0.364 & 0.297 & 0.916 & 0.833 & 0.751 \\
Qwen3.5-4B & Prompt tuning & 0.578 & 0.156 & 0.383 & 0.903 & 0.807 & 0.751 & 0.670 & 0.347 & 0.367 & 0.898 & 0.797 & 0.642 \\
Qwen3.5-4B & Neologism & 0.595 & 0.202 & 0.480 & 0.952 & 0.903 & 0.821 & 0.742 & 0.486 & 0.500 & 0.950 & 0.900 & 0.763 \\
Qwen3.5-4B & LoRA & 0.659 & 0.376 & 0.567 & 0.997 & 0.993 & 0.821 & 0.792 & 0.595 & 0.567 & 0.995 & 0.990 & 0.740 \\
Qwen3.5-2B & No request & 0.529 & 0.087 & 0.317 & 0.835 & 0.670 & 0.890 & 0.564 & 0.127 & 0.320 & 0.840 & 0.680 & 0.815 \\
Qwen3.5-2B & Instruction & 0.549 & 0.110 & 0.323 & 0.880 & 0.760 & 0.896 & 0.584 & 0.173 & 0.317 & 0.880 & 0.760 & 0.838 \\
Qwen3.5-2B & Prompt tuning & 0.560 & 0.133 & 0.397 & 0.932 & 0.863 & 0.867 & 0.601 & 0.214 & 0.387 & 0.928 & 0.857 & 0.809 \\
Qwen3.5-2B & Neologism & 0.567 & 0.145 & 0.367 & 0.925 & 0.850 & 0.896 & 0.641 & 0.295 & 0.367 & 0.921 & 0.843 & 0.757 \\
Qwen3.5-2B & LoRA & 0.659 & 0.353 & 0.543 & 0.998 & 0.997 & 0.896 & 0.763 & 0.538 & 0.543 & 0.998 & 0.997 & 0.827 \\
\bottomrule
\end{tabular}
}
\end{table}

\section{Result Verification}
\label{app:validity}

All reported tables are generated from the canonical item-level result files. We verify that candidate sets remain fixed across paired contexts, that each intervention preserves the intended conflict, and that the reported aggregate scores reproduce the item-level predictions.

\end{document}

%% file: math_commands.tex
\usepackage{amsmath,amsfonts,bm}

\def\eqref#1{equation~\ref{#1}}

\def\1{\bm{1}}

\DeclareMathAlphabet{\mathsfit}{\encodingdefault}{\sfdefault}{m}{sl}
\SetMathAlphabet{\mathsfit}{bold}{\encodingdefault}{\sfdefault}{bx}{n}

